\pdfoutput=1
\documentclass[11pt]{article}

\usepackage[final]{acl}

\usepackage{times}
\usepackage{latexsym}
\usepackage{booktabs} % For professional looking tables
\usepackage{amsmath}
\usepackage{amssymb}
\usepackage{multirow} % For multi-row span in tables
\usepackage{subcaption}
\usepackage{makecell}

\newcommand{\rcomment}[1]{{\color{red} #1}}

\newcommand{\mycomment}[1]{}

\usepackage[T1]{fontenc}
\usepackage[utf8]{inputenc}

\usepackage{microtype}

\usepackage{inconsolata}

\usepackage{graphicx}

\title{Ignore the KL Penalty! Boosting Exploration on Critical Tokens to Enhance RL Fine-Tuning}
\author{
    \begin{minipage}[t]{\textwidth}
        \centering
        Jean Vassoyan\textnormal{\textsuperscript{1,2}} \hspace{2em} 
        Nathanaël Beau\textnormal{\textsuperscript{2,3}} \hspace{2em}
        Roman Plaud\textnormal{\textsuperscript{2,4}} \hspace{2em}
        \\
        \vspace{0.5em} % Adjust the space here as needed
        \textnormal{\textsuperscript{1}Université Paris-Saclay, CNRS, ENS Paris-Saclay, Centre Borelli, France} \\
        \textnormal{\textsuperscript{2}onepoint, France} \hspace{2em}
        \textnormal{\textsuperscript{3}Université de Paris, LLF, CNRS, France} \\
        \textnormal{\textsuperscript{4} Institut Polytechnique de Paris }
        \\
        \vspace{0.3em} % Adjust the space here as needed
        {\normalsize
            \textnormal{\texttt{jean.vassoyan@ens-paris-saclay.fr}} \\  % Optional line for emails
            \textnormal{\texttt{nathanael.beau.gs@gmail.com}} \\  % Optional line for emails
            \textnormal{\texttt{roman.plaud@telecom-paris.fr}}
        }
    \end{minipage}
}

\begin{document}
\maketitle
\begin{abstract}

% Fine-tuning pre-trained LLMs with reinforcement learning helps optimize task-specific objectives, but excessive deviation from the pre-trained model risks degrading foundational capabilities, while insufficient exploration limits discovering better strategies. 
% However, exploration with LLMs is difficult because it's hard to keep the policy close to the pre-trained model while discovering new solutions.
% However, exploration with LLMs is difficult as a balance must be found between discovering new solutions and staying close enough to the pre-trained model, so as not to degrade foundational capabilities.
% We demonstrate the influence of varying pre-training levels on exploration performance and propose a modification to the KL divergence penalty to better balance model proximity and exploration.
% We study the influence of varying pre-training levels on exploration and propose a simple modification to the KL divergence penalty to better balance the proximity with the pre-trained model.
The ability to achieve long-term goals is a key challenge in the current development of large language models (LLMs).
To address this, pre-trained LLMs can be fine-tuned with reinforcement learning (RL) to explore solutions that optimize a given goal.
However, exploration with LLMs is difficult, as a balance has to be struck between discovering new solutions and staying close enough to the pre-trained model, so as not to degrade basic capabilities.
This is typically controlled with a Kullback-Leibler (KL) penalty.
In this paper, we investigate the exploration dynamics of a small language model on a simple arithmetic task. %, under varying degrees of pre-training.
We show how varying degrees of pre-training influence exploration and demonstrate the importance of ``critical tokens'' which have a dramatic impact on the final outcome.
% We show the importance of ``critical tokens'' that have a dramatic impact on model performance and propose a simple modification to the KL penalty that favors exploration on these tokens.
% This paper examines the ability of a small language model to explore out-of-distribution, on a simple arithmetic task.
% We compare exploration dynamics after varying degrees of pre-training and show the existence of ``critical tokens'' that have a dramatic impact on the model performance.
%As a result, we introduce a simple modification to the Kullback-Leibler penalty, that more effectively balances proximity to the pre-trained model.
Consequently, we introduce a simple modification to the KL penalty that favors exploration on critical tokens, increasing the efficiency of the RL fine-tuning stage. \footnote{
Our code and experiments are publicly available at: \url{https://github.com/jvasso/llm-rl-arithmetic}.
}
% We demonstrate how different levels of pre-training influence exploration dynamics and propose a simple modification to the KL divergence penalty, which more effectively balances proximity to the pre-trained model.
% nous mettons aussi en evidence l'existence de critical tokens, qui correspondent à des points charnière de decision sur les taches out of distribution

\end{abstract}

%TODO: remplacer "accuracy" par "pass@1"

\section{Introduction}

% In recent years, expectations on large language models (LLMs) have evolved, moving from a short-term Q\&A paradigm to a task paradigm, where the model is intended to achieve long-term goals \rcomment{<refs ici>}.
In recent years, expectations on large language models (LLMs) have evolved, viewing them more and more as agents intended to achieve long-term goals \citep{wei2023chainofthoughtpromptingelicitsreasoning, llmlongtermgoal, havrilla2024teachinglargelanguagemodels}.
In particular, a number of research studies have found that LLMs can learn to achieve long-term objectives when fine-tuned with Reinforcement Learning (RL), even with a sparse success/failure signal \citep{meta2022human, zelikman2024quiet, havrilla2024teachinglargelanguagemodels, guo2025deepseek}.
In such setting, a pre-trained language model is typically used as a policy to explore solutions within a text-generation task.
Pre-training plays an ambivalent role in guiding exploration: on the one hand, the policy should not deviate too far from the pre-trained model in order to maintain basic capabilities (like language structure) -- this is why a KL-divergence penalty is typically added to the loss \cite{ziegler2019fine}.
On the other hand, staying too close to the pre-trained model can significantly hinder its potential for exploration.
On this matter \cite{havrilla2024teachinglargelanguagemodels} have demonstrated that LLM agents typically fail to explore beyond solutions produced by the pre-trained models.
% We hypothesize that a better control of the balance between old and new policies can enhance the model's exploration capabilities, especially as the distribution shift increases between pre-training and fine-tuning.
We hypothesize that more precisely balancing the trade-off between old and new policies can improve the model’s exploration capabilities, especially as the distribution shift increases between pre-training and fine-tuning.

\begin{figure}
    \centering
    \includegraphics[width=\linewidth]{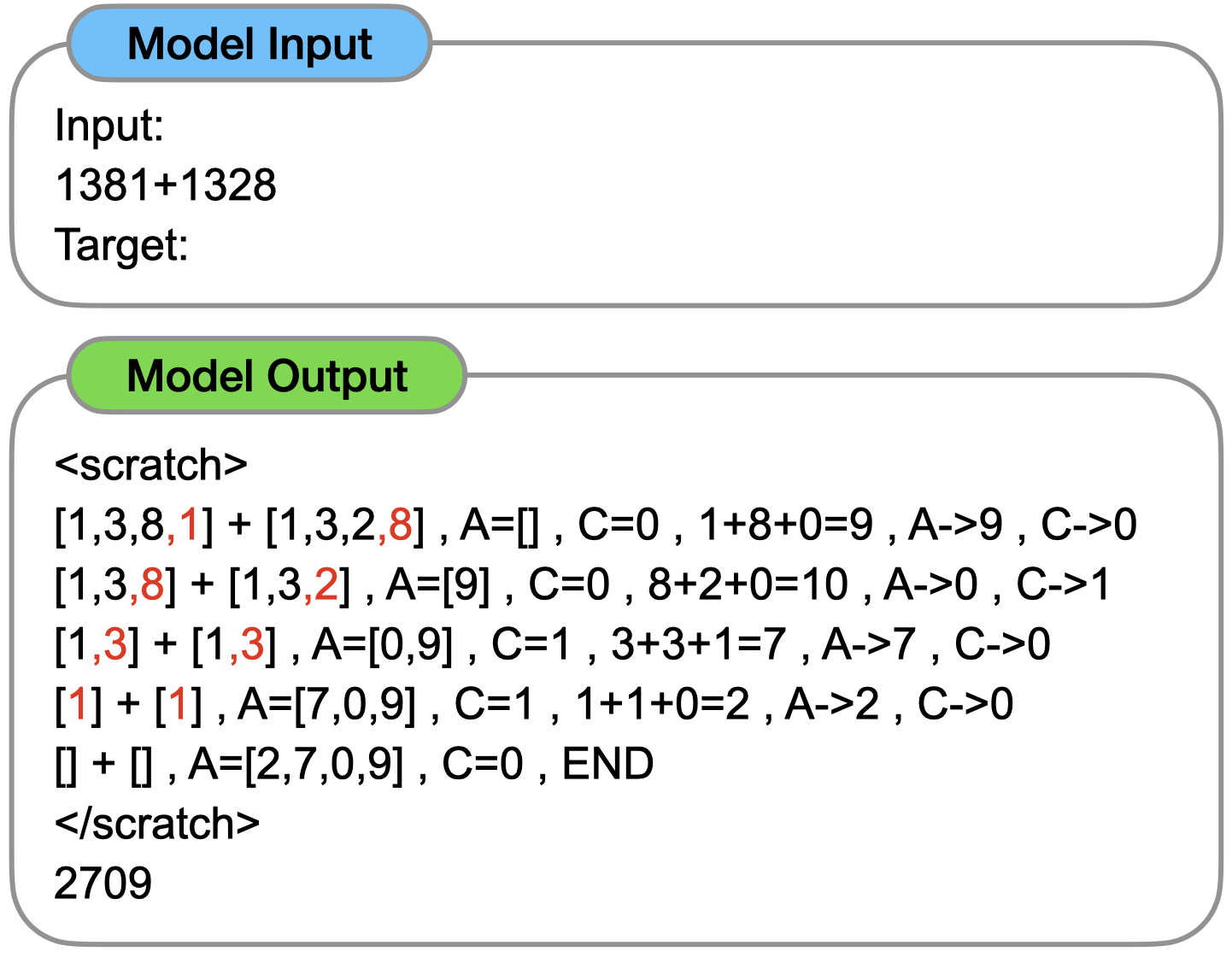}
\caption{Illustration of the addition task with scratchpad, for a model pre-trained on numbers up to 3 digits.
The highlighted critical tokens are decision points where the model tends to make mistakes, mainly because it is tempted to process the number as if it were shorter. This occurs when the model is faced with a number that is longer than those encountered during the pre-training stage (here, 4 digits instead of 3).
% the highlighted critical tokens are key decision points that slightly differ from those encountered in pre-training
}
\label{fig:scratchpad}
\end{figure}

This article examines how varying levels of pre-training affect language model performance in a task requiring some level of exploration.
We introduce an experimental setup where the model is first pre-trained on a simple arithmetic task, then fine-tuned with RL on a similar task with a small distribution shift.
We chose the arithmetic task for two main reasons.
%The arithmetic task was chosen for two reasons:
First, prior research highlights the value of studying language models on basic arithmetic problems \cite{GOAT, zhou2024transformersachievelengthgeneralization}, noting challenges in generalizing to novel digit lengths — though these difficulties vary by model type and use of scratchpads \citep{benchmarkllmarithmetic, lee2023teachingarithmeticsmalltransformers}.
Second, this task closely mirrors real-world LLM applications while enabling fine-grained control over the distribution shift between pre-training and RL fine-tuning stages.
% Notably, we find that performance on the RL task with a distribution shift depends on a few critical tokens where the policy must diverge from the pre-trained model’s predictions.
Notably, we find that performance on this RL task is determined by a few critical tokens where the policy must diverge from the pre-trained model’s predictions.
This observation motivated a modification to the original KL penalty, making it more dependent on the pre-trained model's confidence.

Our contribution is two-fold: we first conduct an analysis of the influence of pre-training on a small language model's ability to explore out-of-distribution.
% More precisely, we investigate how pre-training on a greater diversity of operand lengths can affect exploration performance on new operand lengths.
More precisely, we investigate how pre-training with a broader range of operand lengths influences the model's performance on new operand lengths.
Second, we introduce a simple trick that allows to adapt the KL penalty to the token-wise confidence of the pre-trained model.
Our empirical results show that this modification to the KL penalty substantially enhances exploration efficiency.

\section{Related Work}

\paragraph{LLMs and Reasoning}
Recent state-of-the-art LLMs~\citep{llama2,GPT4} have shown strong performance on reasoning tasks across various benchmarks, including mathematics~\citep{cobbe2021trainingverifierssolvemath, hendrycks2021measuringmathematicalproblemsolving} and code~\citep{chen2021evaluatinglargelanguagemodels, Li_2022}. Combining LLMs with prompting strategies like chain-of-thought~\citep{wei2023chainofthoughtpromptingelicitsreasoning} has become a common approach for tackling complex reasoning tasks by guiding the model to break down problems into smaller subproblems.
% maybe talk about llm + agents

% -benchmarks reasoning (math) + CoT, ToT etc 
% -llm + agents ?\\
\paragraph{LLMs and RL}
The integration of LLMs and RL has primarily been driven by Reinforcement Learning from Human Feedback (RLHF) \citep{NIPS2017_d5e2c0ad, ziegler2019fine, NEURIPS2020_1f89885d}, which aligns model outputs with human preferences.
% These methods often rely on the Proximal Policy Optimization (PPO) algorithm \citep{ppo}, typically combined with a KL penalty to prevent excessive deviation from the pre-trained policy \citep{ziegler2019fine}.
However we stress that learning from human preferences is a different framework from the more general one of RL, as the latter focuses on optimizing long-term objectives -- possibly with high level of exploration -- while learning from human preferences can be achieved solely with a fixed dataset.
% Although RLHF is intended to "rank" already existing solutions, RL has also been applied to LLMs in the standard exploration/exploitation framework in tasks such as grounding \citep{yao2020calmexplorelanguagemodels, carta2023groundinglargelanguagemodels}, code generation \citep{NEURIPS2022_8636419d}, and mathematical reasoning \cite{havrilla2024teachinglargelanguagemodels}.
RL has also been applied to LLMs in this more general framework, in tasks such as grounding \citep{yao2020calmexplorelanguagemodels, carta2023groundinglargelanguagemodels}, code generation \citep{NEURIPS2022_8636419d}, and mathematical reasoning \cite{havrilla2024teachinglargelanguagemodels}.
Training LLMs with RL presents challenges due to reward sparsity \citep{cao2024sparserewardsenhancingreinforcement}, credit assignment difficulties in identifying key actions that led to failure \citep{hwang2024selfexploreenhancingmathematicalreasoning}, large state spaces requiring exploration, and unstable training processes. \citet{havrilla2024teachinglargelanguagemodels} have raised concerns about RL algorithms, struggling to explore beyond solutions already produced by supervised fine-tuning (SFT) models.

\paragraph{LLMs and Addition} The addition task remains challenging even for the latest LLMs, which struggle to accurately add large numbers and track digit positions~\citep{wallace-etal-2019-nlp}. Most related studies have focused on supervised learning approaches~\citep{lee2023teachingarithmeticsmalltransformers, mcleish2024transformersarithmeticrightembeddings} and improving positional encoding~\citep{shen2023positionaldescriptionmatterstransformers, kazemnejad2023impactpositionalencodinglength,mcleish2024transformersarithmeticrightembeddings, zhou2024transformersachievelengthgeneralization}. Generalization to unseen lengths is a common evaluation criterion in these studies~\citep{kazemnejad2023impactpositionalencodinglength, xiao2023conditionslengthgeneralizationlearning, zhou2024transformersachievelengthgeneralization}.
Despite the addition task being a reasoning problem with a well-defined long-term reward, no research, to our knowledge, has addressed it using RL with a language model.
The closest work is by \citet{zhang2023chainofthoughtreasoningpolicyimprovement}, who incorporated a self-training loop after the supervised fine-tuning phase.

\section{Problem formulation}

\subsection{Addition as a Markov Decision Process}

We propose to study the performance of a language model on a simple arithmetic task.
The model is prompted to perform the addition of two numbers whose lengths range from $1$ and $N$.
To do this, it has to break down the calculation step by step, following a predefined scratchpad.
In practice, we opted for the scratchpad from \citep{lee2023teachingarithmeticsmalltransformers} with minor modifications (see Figure~\ref{fig:scratchpad}).

This task can be simply expressed as a Markov Decision Process \mbox{$\mathcal{M} = (\mathcal{S},\mathcal{A},\mathcal{T},\mathcal{R})$} where the action space $\mathcal{A}$ is the vocabulary, each state $s_t\in\mathcal{S}$ is the text generated up to $t$ steps, with $s_0$ the initial prompt and $\mathcal{T}$ the (deterministic) transition function that derives directly from the actions taken by the model.
The reward function $\mathcal{R}$ is $0$ all along the episode, and takes the value of $1$ if the final result is correct (0 otherwise).
As in most reinforcement learning problems, the goal is to find a policy \mbox{$\pi :\mathcal{S}\rightarrow \mathcal{A}$} that maximizes the expected return over each episode: \mbox{$\pi^* = \underset{\pi}{\arg \max }\ \mathbb{E} \left[\  \sum_{t=0}^{T-1} \mathcal{R}(\mathbf{s}_t) \ \right]$}.
We directly take the language model, denoted $\pi_\theta$, as the policy.

\subsection{Experimental setting}
\label{sec:exp_setting}
Our experimental pipeline consists in pre-training the language model on number lengths ranging from $1$ to $N$, then fine-tuning it with RL on number lengths $N+1$ or $N+2$.

% During the pre-training stage, following the methodology from \citep{teachingarithtosmalltransformer}, we trained the language model from scratch, in a supervised-learning manner, on a dataset of scratchpad examples.
% In the pre-training phase, we followed the methodology outlined by \citet{teachingarithtosmalltransformer}, training the language model from scratch using supervised learning on a dataset of scratchpad examples. 
% The dataset was generated in such a way that each number length is equally represented from $1$ to $N$.
% The dataset was constructed to ensure equal representation of number lengths ranging from $1$ to $N$. 
% This process leads to a pre-trained language model -- denoted $\pi_{\theta_{\text{old}}}$ -- that performs very well on number lengths up to $N$, but also exhibits some level of generalization over $N+1$, $N+2$, $N+3$, especially for larger values of $N$.
%As a result, the pre-trained language model, denoted $\pi_{\theta_{\text{old}}}$, achieved strong performance on number lengths up to $N$.
% Notably, the model also demonstrated a degree of generalization to longer sequences beyond the training range, specifically for lengths $N+1$, $N+2$, and $N+3$, with this effect being more pronounced for larger values of $N$.
%Details on evaluation methodology can be found in Appendix~\ref{app:evaluation-methodology}.
In the pre-training phase, we followed the approach from \citet{lee2023teachingarithmeticsmalltransformers}, training the language model from scratch using supervised learning on a scratchpad dataset.
The dataset was balanced across number lengths from $1$ to $N$, ensuring uniform representation.
The resulting pre-trained model $\pi_{\theta_{\text{old}}}$ performs well on numbers up to length $N$.
The evaluation was conducted on two setups: fixed digit addition, where both terms had exactly $N$ digits, and varying digit addition, where one term had $N$ digits and the other had fewer. %, to assess generalization.
More details on the evaluation methods are provided in Appendix~\ref{app:evaluation-methodology}.

For the RL fine-tuning stage, we initialized the policy with $\pi_\theta = \pi_{\theta_{\text{old}}}$ and performed training on number lengths $N+1$ or $N+2$.
This corresponds to an ``out-of-distribution'' scenario that the model cannot reliably handle without further training.
As a result, the only way for the model to succeed in this new task is to explore, so as to identify the errors it makes in the scratchpad and correct them.

\subsection{Critical tokens}

% Without loss of generality, we propose to define as "critical token" any step in the generation process that meets these two criteria:
% \begin{itemize}
%     \item it is decisive for the rest of the answer: if the model is wrong about this token, the overall answer will most certainly be wrong (the model fails to "correct" itself);
%     \item the model shows substantially more uncertainty on these tokens than on the others.
% \end{itemize}
% An interesting outcome of our experiments is the emergence of a small number of tokens that have a dramatic impact on the final result.
A notable finding from our experiments is the emergence of a small subset of tokens that significantly influence the final outcome.
We refer to these as ``critical tokens'' and define them as follows.
Within the output generated by a language model, a ``critical token'' is a token that satisfies both of these criteria:
\begin{itemize}
    \item it is decisive for the rest of the answer: if the model is wrong about this token, the final answer will most likely be wrong (the model fails to correct itself);
    \item the pre-trained model shows substantially more uncertainty on these tokens than on the rest of the output.
\end{itemize}
In our experiments, these tokens arise when the model has to act in a different way from that encountered during pre-training (out-of-distribution decision making).
More precisely, if the model is pre-trained on numbers up to $N$ digits, critical tokens occur in the decomposition stages that process the (N+1)-th or (N+2)-th digit (highlighted in Figure \ref{fig:scratchpad}).
Regarding the first criterion, we found that whenever these tokens are generated incorrectly, the model inevitably produces the wrong answer.
% Regarding the first criterion, our empirical results show that whenever these tokens are generated incorrectly, the model inevitably produces the wrong answer.
As for the second criterion, we carried out a quantitative analysis comparing the model's certainty on these tokens against the others.
More precisely, for each token, we measured the quantity $\Delta\widehat{J}_{\theta_{\text{old}}}(s)$, defined as the difference between the certainty on this token and the mean certainty on the others.
% $\Delta\widehat{J}_{\theta_{\text{old}}}(s)$ is expected to be negative on critical tokens and positive on the others.
The results, reported in Table \ref{tab:uncertainty}, show a significant gap in certainty between the critical tokens and the rest of the output.
% and shows significantly more uncertainty on them than on the rest of the output (see Table \ref{tab:uncertainty}).
More details on these critical tokens and their location in the scratchpad are provided in Appendix~\ref{app:critical_tokens}.

\begin{table}[t]
\centering
\scalebox{1}{
\begin{tabular}{c|cc}
\toprule
  & \makecell{$\Delta\widehat{J}_{\theta_{\text{old}}}(s)$ \\ critical}  & \makecell{$\Delta\widehat{J}_{\theta_{\text{old}}}(s)$ \\ non-critical (min.)}  \\ \midrule
$N=3$  & -0.33 $\pm$ 0.01 & 0.0012 $\pm$ 0.0001 \\
$N=5$  &  -0.21 $\pm$ 0.18 & 0.0002 $\pm$ 0.0001 \\
$N=7$ & -0.13 $\pm$ 0.04 & 0.0004 $\pm$ 0.0001\\ 
\bottomrule
\end{tabular}}
\caption{Comparison of the quantity $\Delta\widehat{J}_{\theta_{\text{old}}}(s)$ for critical and non-critical tokens, averaged over 50 generations. This shows the model's high level of uncertainty on critical tokens.
%Difference between the certainty of a given token and the average certainty of all other tokens. Left column is the quantity for the critical token while right column is the mean quantity for all other tokens.
}
\label{tab:uncertainty}
\end{table}

\section{Prioritized KL penalty}
\label{sec:weightedKLtrick}
When fine-tuning a language model with RL, a Kullback-Leibler (KL) penalty term is usually added to the loss to avoid deviating too far from the pre-trained model: $\mathcal{L} = \mathcal{L}_{\text{RL}} + \alpha \mathcal{L}_{\text{KL}}$ where $\mathcal{L}_{\text{KL}} = \mathbb{E}_{s,a \sim \pi_{\theta}} \left[ \log \frac{ \pi_{\theta}(a|s)}{\pi_{\theta_{\text{old}}}(a|s)} \right]$ and $\pi_{\theta_{\text{old}}}$ is the pre-trained model.
As a result, the target policy $\pi_\theta$ is encouraged to approach the predictions of $\pi_{\theta_{\text{old}}}$ on each state-action pair.
We argue that this penalty term could lead to more efficient exploration out of distribution if each state-action term was weighted by the certainty on the old policy predictions:
\begin{equation}\label{eq:weighted_kl}
    \widetilde{\mathcal{L}}_{\text{KL}} = \mathbb{E}_{s,a \sim \pi_{\theta}} \left[ \widehat{J}_{\theta_{\text{old}}}(s)^{\beta} . \log \frac{ \pi_{\theta}(a|s)}{\pi_{\theta_{\text{old}}}(a|s)} \right]
\end{equation}
where $\widehat{J}_{\theta_{\text{old}}}(s)$ estimates the certainty of the pre-trained model in state $s$ and $\beta$ is a hyperparameter.
This quantity can be taken as the normalized negentropy \citep{brillouin1953negentropy}, which is negatively correlated with entropy: $J = \frac{H_{\text{max}}-H}{H_{\text{max}}}$.
In an ideal scenario, one would not only account for the data uncertainty but also for the model uncertainty, for example leveraging a bayesian approach\footnote{
% \begin{equation}
In a bayesian approach, one would provide an estimate of $J$ not only based on data uncertainty but also on model uncertainty: $J(s) = J \left[ \int_{\theta_{\text{old}}} \pi_{\theta_{\text{old}}}(\cdot|s) p(\theta_{\text{old}} | \mathcal{D}_{\text{pretrain}}) d\theta_{\text{old}} \right]$
% \end{equation}
.}.
However, since our framework falls within a context where the pre-trained model is given and fixed, we deliberately settle for an approximation that does not take into account this type of uncertainty.
Our final estimate is as follows:
% We propose a slight change to this term in order to better account to
\begin{equation}
    \widehat{J}_{\theta_{\text{old}}}(s) = \frac{H_{\text{max}}-H(\pi_{\theta_{\text{old}}}(\cdot|s))}{H_{\text{max}}}
\end{equation}

Our results in the next section show that, although the penalty term from Equation~\ref{eq:weighted_kl} does not address crucial aspects such as model overconfidence, it outperforms the standard KL penalty in our experimental setting.

\section{Experimental results}

\begin{figure}[t]
\begin{center}
\scalebox{0.48}{
\includegraphics[width=\textwidth]{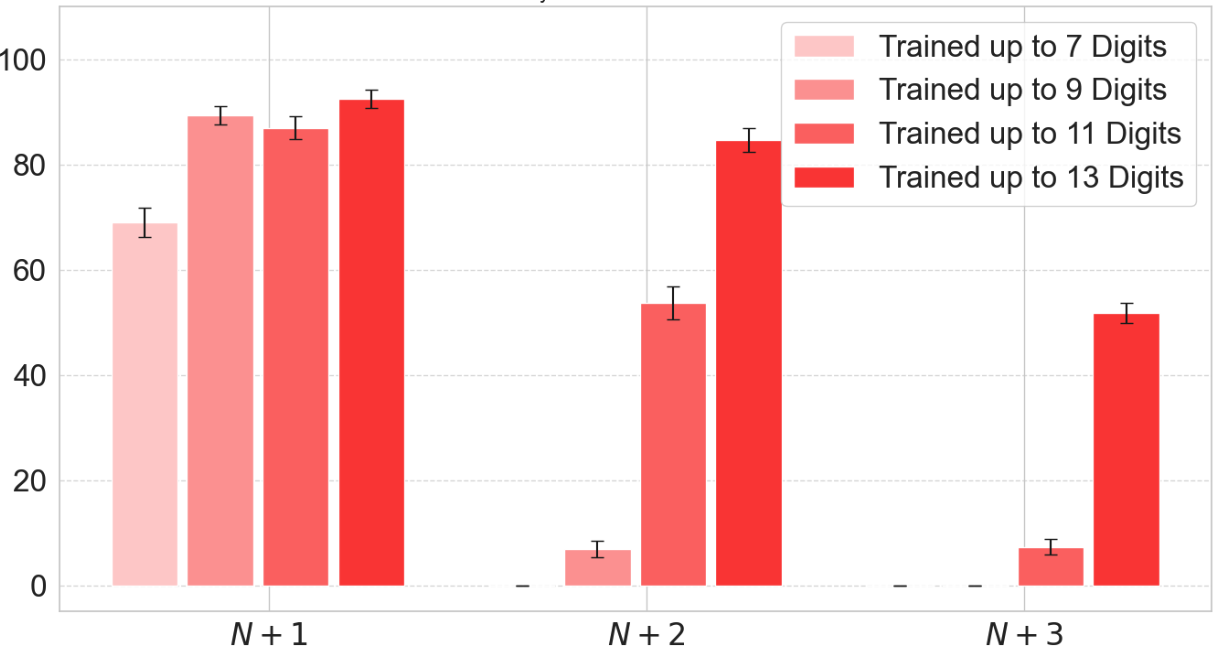}}
\end{center}
\caption{\label{fig:pretresult}Model accuracy on addition tasks for models trained on numbers up to digit lengths $N=7, 9, 11, 13$.
%The darker the color, the more digits the model was trained on.
Results are shown for varying digit evaluation. Error bars indicate 95\% confidence intervals. Full detailed results are provided in Appendix~\ref{app:detailed-pretraining-results}.}
\end{figure}

\subsection{Training Details}
All experiments were carried out with the GPT-2 language model \cite{GPT2}.
%serves as the base model for our policy, selected to keep training compute and inference time reasonable.
% A character-level tokenizer with a 96-size vocabulary ensures proper representation of digits, facilitating addition tasks \citep{wallace-etal-2019-nlp}.
A character-level tokenizer was used to ensure proper representation of digits, facilitating addition tasks \citep{wallace-etal-2019-nlp}.
The resulting model had 85M parameters.
The reinforcement learning experiments were carried out with A2C \citep{a2c}.
We chose this algorithm because it is both simple and efficient, with few hyperparameters, making it more suitable for our comparison purposes.
When applicable, the computation of the KL divergence was approximated with the estimator from \citet{schulman2023approximating}: $\mathrm{KL}[q, p] \approx \frac{1}{2}(\log p(x)-\log q(x))^2$.
The hyperparameters used for each experiment are provided in Appendix~\ref{app:experiments_details}.

% \subsection{Evaluation of Pretraining Effectiveness}
\subsection{Comparison of varying levels of pre-training}\label{subsec:compare_pretrain}

Before the application of any fine-tuning with RL,
we show in Figure \ref{fig:pretresult} that increasing the number $N$ of digits during the pre-training stage improves generalization on addition tasks with larger numbers of digits.
The same trend holds for equal-length addition evaluations, where models trained on larger $N$ demonstrate better generalization.
Detailed results on each task are provided in Appendix~\ref{app:detailed-pretraining-results}.

% \rcomment{J'enleverais cette phrase du core paper -> (}
% In particular, for identical digit addition tasks, accuracy reaches 67.7\% on $N+2$ with larger $N$. For varying digit addition, generalization is stronger, achieving 84.7\% on $N+2$ and 51.8\% on $N+3$ when pre-trained on $N=13$, reflecting closer alignment with the pre-training data distribution.
% \rcomment{)}

\begin{figure}
    \centering
    \includegraphics[width=0.99\linewidth]{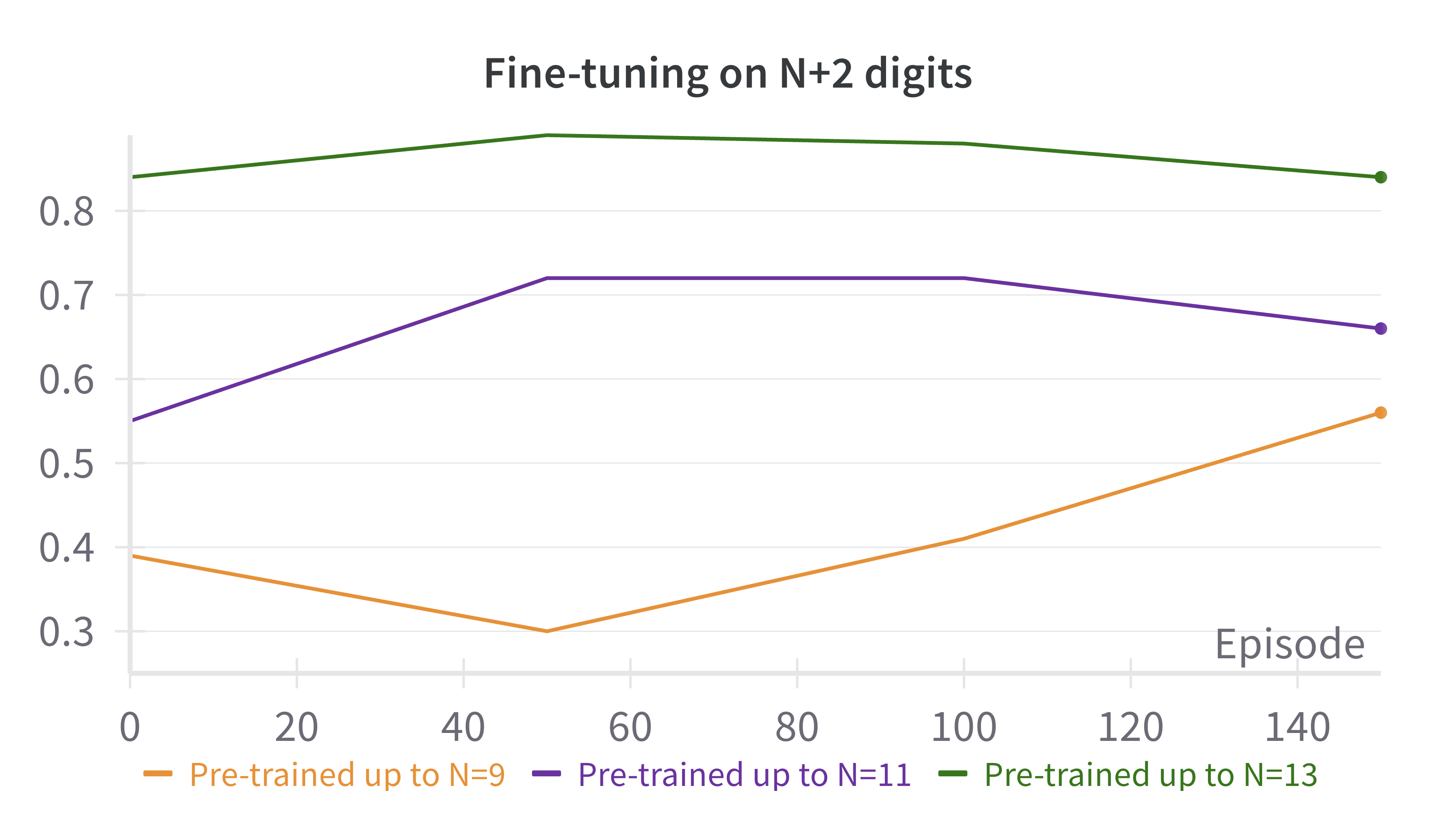}
    \caption{Learning curves of multiple models pre-trained up to $N$, fine-tuned with RL on $N+2$.}
    \label{fig:rl_compare_pretrain}
\end{figure}

In another experiment, we fine-tuned each pre-trained model with RL and examined their performance on additions with $N+1$ digits.
The results are reported in Figure~\ref{fig:rl_compare_pretrain}.
% Surprisingly, the models pre-trained on a higher number of digits $N$, although initially more efficient, failed to maintain their lead throughout the exploration phase, and were eventually overtaken by the models pre-trained on smaller values of $N$.
% Surprisingly, the models pre-trained on a higher number of digits, although more efficient initially, tend to stagnate during the exploration stage.
Interestingly, the models pre-trained on more digits — despite being initially more effective — tend to plateau during the exploration phase.
% It is possible that making fewer mistakes at the beginning does not encourage exploration.
One possible explanation is that making fewer early mistakes reduces the incentive to explore.
% Moreover, a qualitative analysis of the scratchpads generated by these models revealed that the errors they make are of a less generic type (mainly copying or token duplication errors) than the ones related to the critical tokens.
Moreover, a qualitative analysis of the scratchpads generated by these models revealed that the errors they make (mostly copying or token-duplication issues) are less generic than those related to critical tokens.
Correcting such errors may require substantially more training steps.
% It is true that the models pre-trained on larger values of make fewer errors from the very beginning, and that this doesn't necessarily encourage exploration. Actually, a qualitative analysis of the scratchpads generated by these models has shown us that the errors they make are of a less generic type (mainly recopying or token duplication errors) than the ones related to the critical tokens. Consequently, it is possible that this type of error requires a much larger number of training steps to be corrected.
% do not manage to correct their mistakes faster than the ones pre-trained on smaller values of $N$.

% We first conducted an analysis of each model performance out of distribution, without any further training.
% We hypothesize that this might be related to the fact that episodes are much longer with higher values of N.
% There are two ways to interpret this result: First, a model pre-trained on higher value of N aquires a higher level of generalization. Second, the distribution shift is actually smaller for such model

\subsection{Impact of the prioritized KL penalty}\label{subsec:compare_kl_trick}

\begin{figure}
     \centering
     \begin{subfigure}[b]{1\linewidth}
         \centering
         \includegraphics[width=\textwidth]{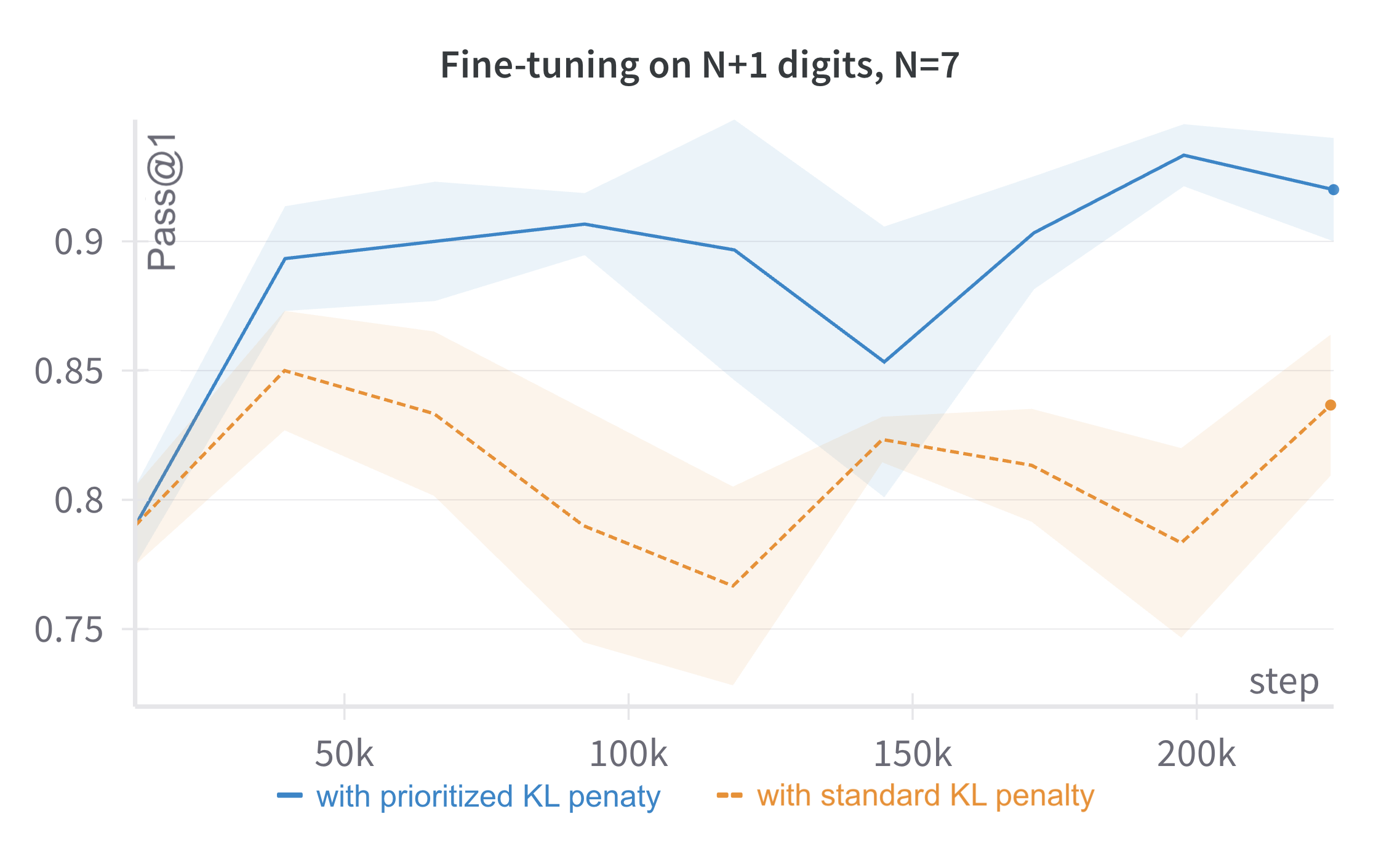}
     \end{subfigure}
     \begin{subfigure}[b]{0.49\linewidth}
         \centering
         \includegraphics[width=\textwidth]{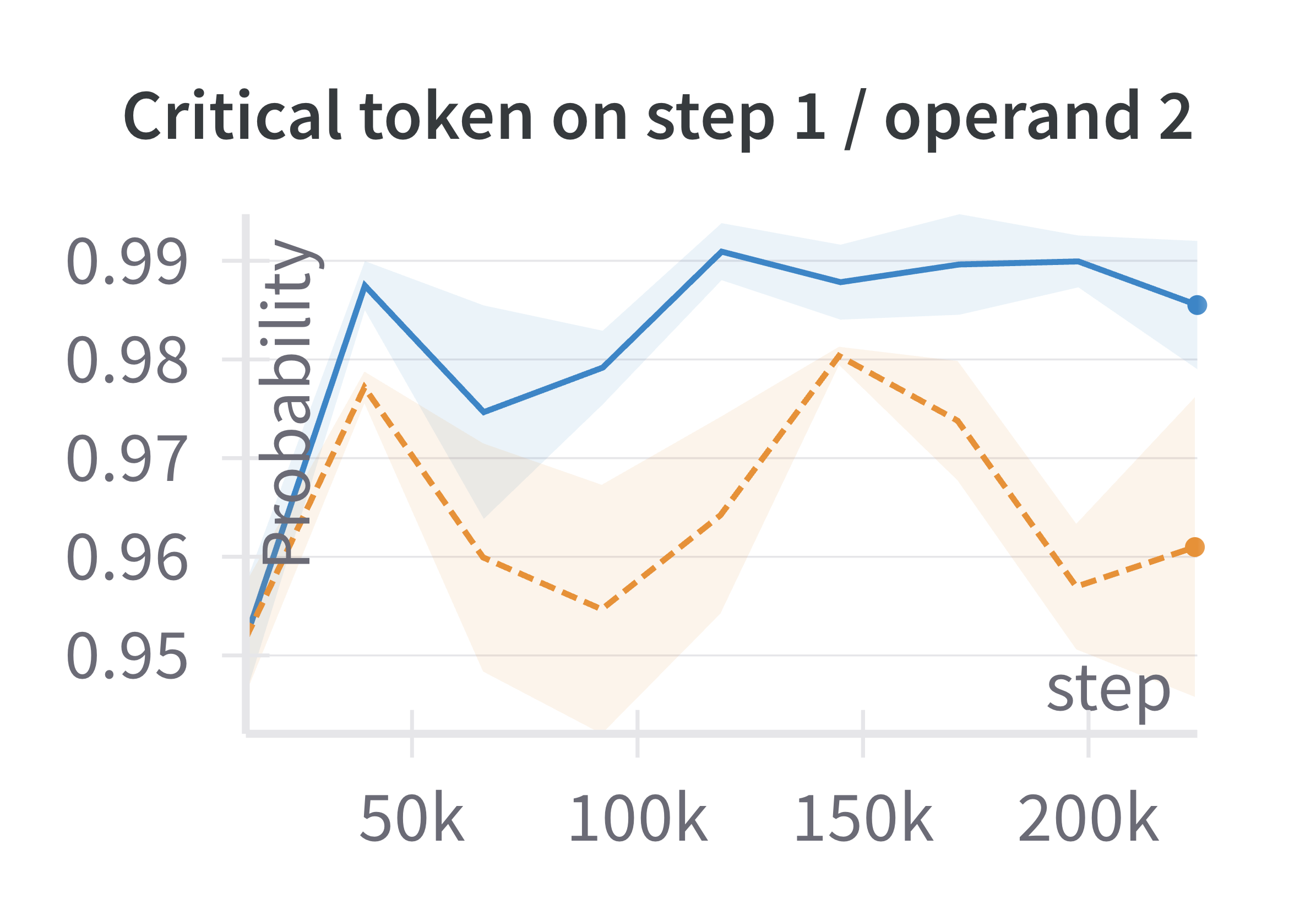}
     \end{subfigure}
     \hfill
     \begin{subfigure}[b]{0.49\linewidth}
         \centering
         \includegraphics[width=\textwidth]{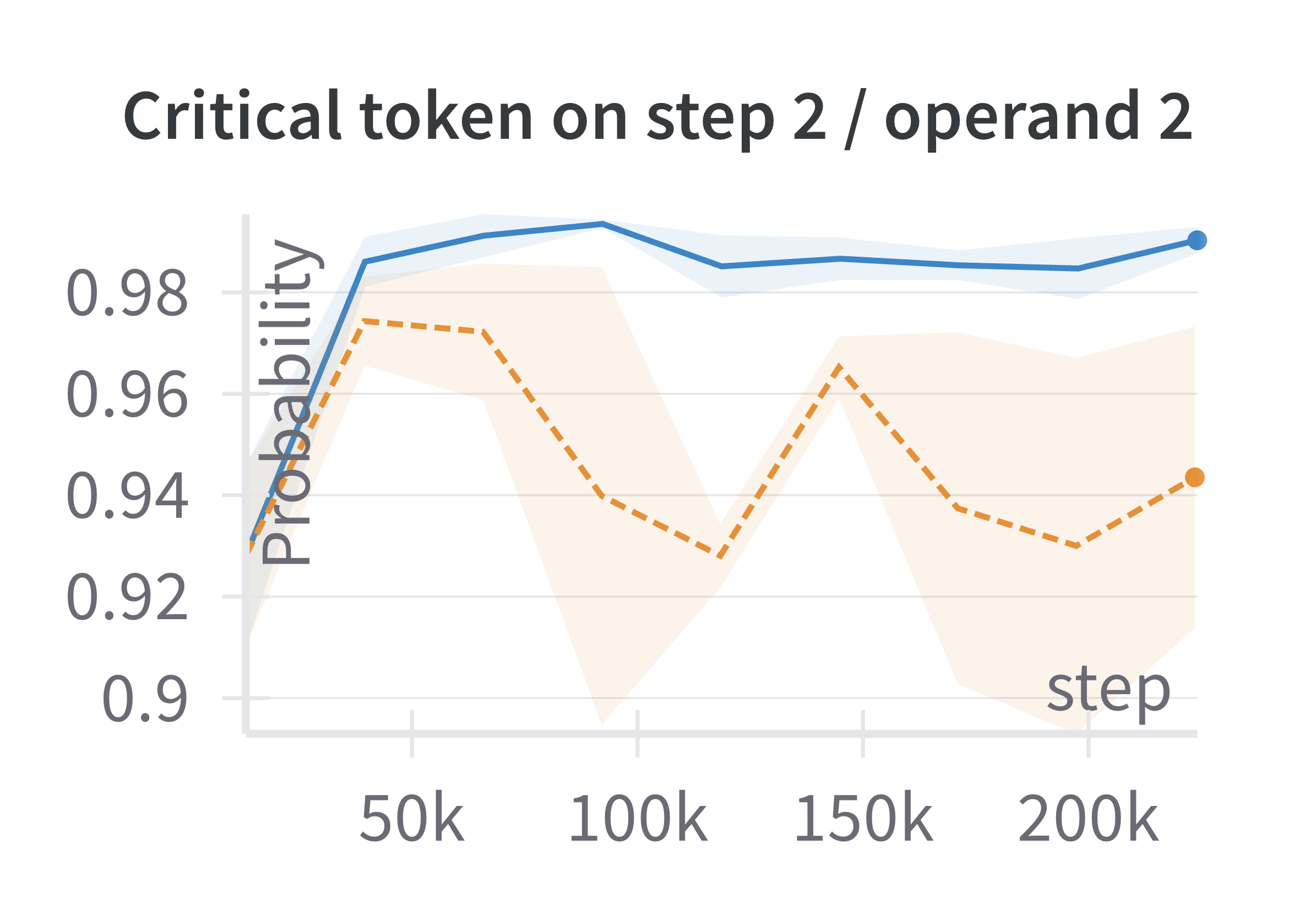}
     \end{subfigure}
        \caption{Top: Learning curves of a model fine-tuned with RL on N+1=8 digits. Bottom: Probability of making the right prediction on two critical tokens. Results on more critical tokens are provided in Appendix~\ref{app:rl_experiments}.}
    \label{fig:results_kl_trick_7d}
\end{figure}

To assess the effectiveness of the prioritized KL penalty, we conducted an experiment where a pre-trained model was fine-tuned with RL using this trick and compared it against a fine-tuning with the standard KL penalty.
We chose to run this experiment on $N=7$ digits as this is the first value of $N$ for which generalization capabilities emerge after pre-training.
The resulting learning curves are provided in Figure \ref{fig:results_kl_trick_7d}.
From these results, one can notice that the model that benefited from the prioritized KL penalty significantly outperformed the other one.
We also provide, on the same figure, some curves depicting the probability of making the right prediction on two critical tokens.
% Interestingly, the first model managed increase the probability of the right prediction and keep it high on the long term, while its counterpart regularly returned to its initial value, probably due to the influence of the standard KL divergence.
Notably, the first model consistently increased and maintained a high probability of correct predictions over the long term, whereas the other one frequently reverted to its initial probability levels, likely due to the effects of the standard KL divergence.
In Appendix~\ref{app:ablation_study}, we test multiple orders of magnitude for the value of the exponent $\beta$ and show that the performance gain provided by the prioritized KL penalty is robust over a wide range of values.

% \begin{figure}
%     \centering
%     \includegraphics[width=0.99\linewidth]{figures/results_kl_trick_7d.png}
%     \caption{Learning curves of a model fine-tuned with RL on N+1=8 digits, with or without the weighted KL divergence trick.}
%     \label{fig:results_kl_trick_7d}
% \end{figure}

% \begin{figure}
%     \centering
%     \includegraphics[width=0.99\linewidth]{figures/critical_tokens_probas/ct_s2o2.png}
%     \caption{Probability of making the right prediction on the critical token located in step 2, operand 2.}
%     \label{fig:results_kl_trick_7d}
% \end{figure}

\section{Conclusion}

In this paper, we studied the performance of a language model pre-trained with supervised learning and fine-tuned with RL on a simple arithmetic task.
% We showed how this experimental setting allowed to identidy 
We showed that this experimental setting allowed to identify a new error mode -- critical tokens -- featuring decisions out of the pre-training data distribution.
Therefore, we proposed a simple trick -- the prioritized KL penalty -- allowing to boost exploration on these tokens during the RL fine-tuning stage.
In future work, we will try to extend the analysis of critical tokens to broader domains and examine the possible application of the prioritized KL penalty to more standard LLM problems.

% Futur work: pousser plus loin l'analyse des critical tokens à des domaines plus larges et examiner l'application éventuelle de la weighted KL divergence à des problèmes plus standards

\section{Limitations}

The main limitation of our study relates to the restricted experimental setup, which limits the scope of the results.
Our experiments were carried out with a small language model, GPT-2, with much less capabilities than the newer, bigger models.
As a result, the task is far less challenging than the benchmarks usually used to evaluate LLMs.
% Second it was done on a very simple task, which is not the kind of task on which one would normally use a language model.
However, this simplicity is also a strength as it allows to study the behavior of the model in a very flexible environment, with more control over the distribution shift.
Moreover, the use of a formatted scratchpad for each answer allowed to easily run statistics about the model behaviour on critical tokens.

\section*{Acknowledgment}

We warmly thank Matthieu Labeau for reviewing an earlier version of this paper and offering valuable feedback.
We also thank Nicolas Vayatis, Pirmin Lemberger, Antoine Saillenfest and Ben Kabongo for insightful discussions about this work.
This work was granted access to the HPC resources of IDRIS under the allocation 2024-TMP32592 made by GENCI.

% \section*{Acknowledgments}

\bibliography{acl_latex}

\newpage
\appendix

\section{Evaluation Methodology}
\label{app:evaluation-methodology}

The evaluation methodology assesses the performance of models pre-trained on digit addition tasks, following the framework of \citet{lee2023teachingarithmeticsmalltransformers}.
Each model, denoted as $\pi_{\theta_{\text{old}}}$, is pre-trained using supervised learning on addition tasks involving up to $N$ digits. The evaluation is conducted under two scenarios:
\begin{enumerate}
    \item \textbf{Identical Digit Addition:} Both terms in the addition consist of exactly $N$ digits (i.e., $N + N$ digit addition).
    
    \item \textbf{Varying Digit Addition:} The model is tested on addition tasks where the number of digits in the two terms varies (i.e., $N + M$ digit addition, where $M \leq N$). The pairs of numbers with different digit counts are sampled to ensure a broader range of difficulty.
\end{enumerate}

Model outputs are evaluated by comparing the predicted results to the ground truth for each addition. Accuracy is computed as the proportion of correct predictions over the total number of examples.

The evaluation is performed on 1,000 test examples. To account for variability in performance, results include confidence intervals obtained via resampling. 

\section{Critical tokens}\label{app:critical_tokens}

\begin{figure}[ht]
     \centering
     \begin{subfigure}[b]{\columnwidth}
        \centering
        \includegraphics[width=\textwidth]{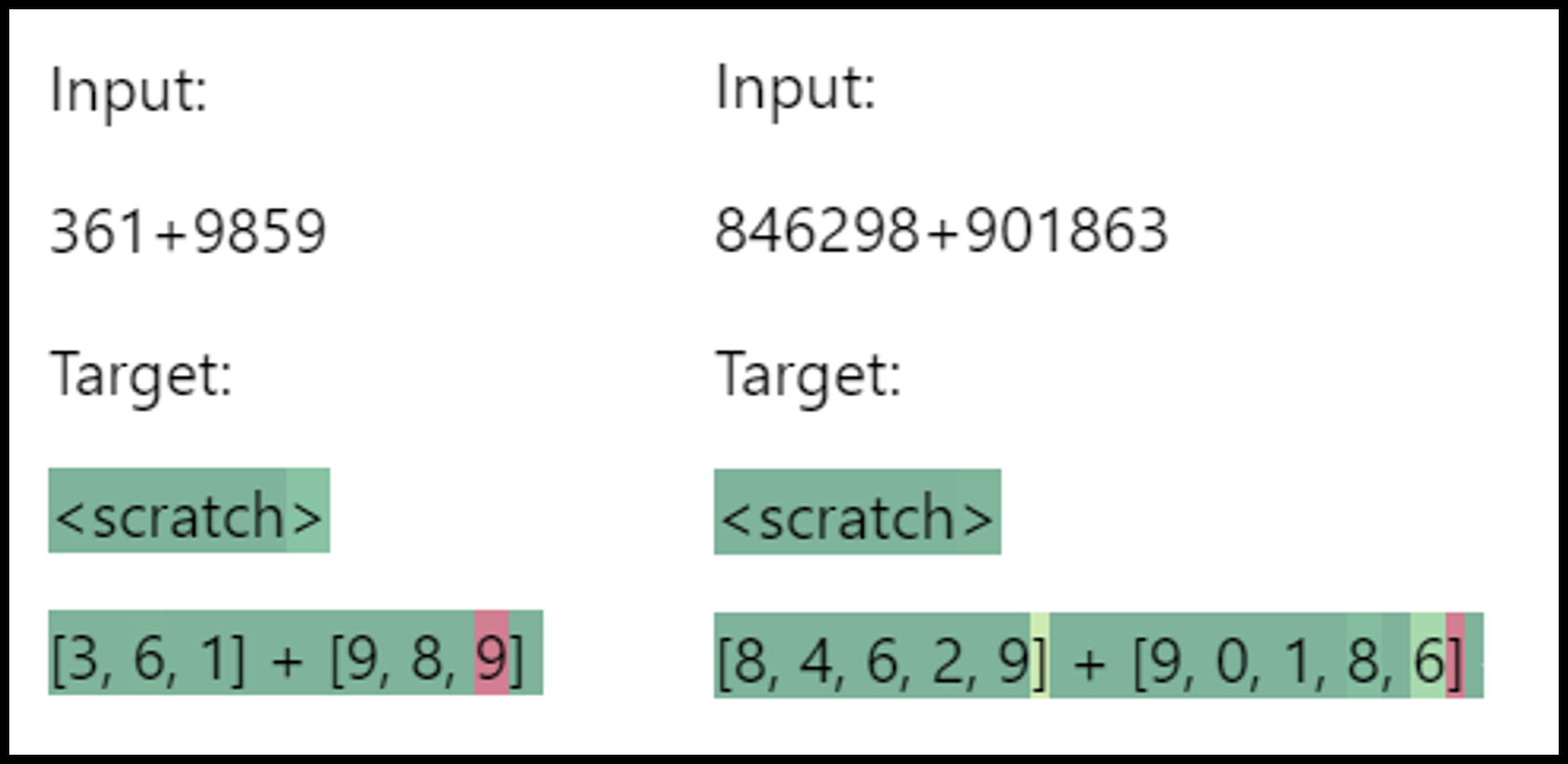}
        \caption{%Failed scratchpad generations, caused by errors on the critical tokens.
        }
        \label{fig:ex_certainty}
     \end{subfigure}
     % \hfill
     \begin{subfigure}[b]{\columnwidth}
        \centering
        \includegraphics[width=\textwidth]{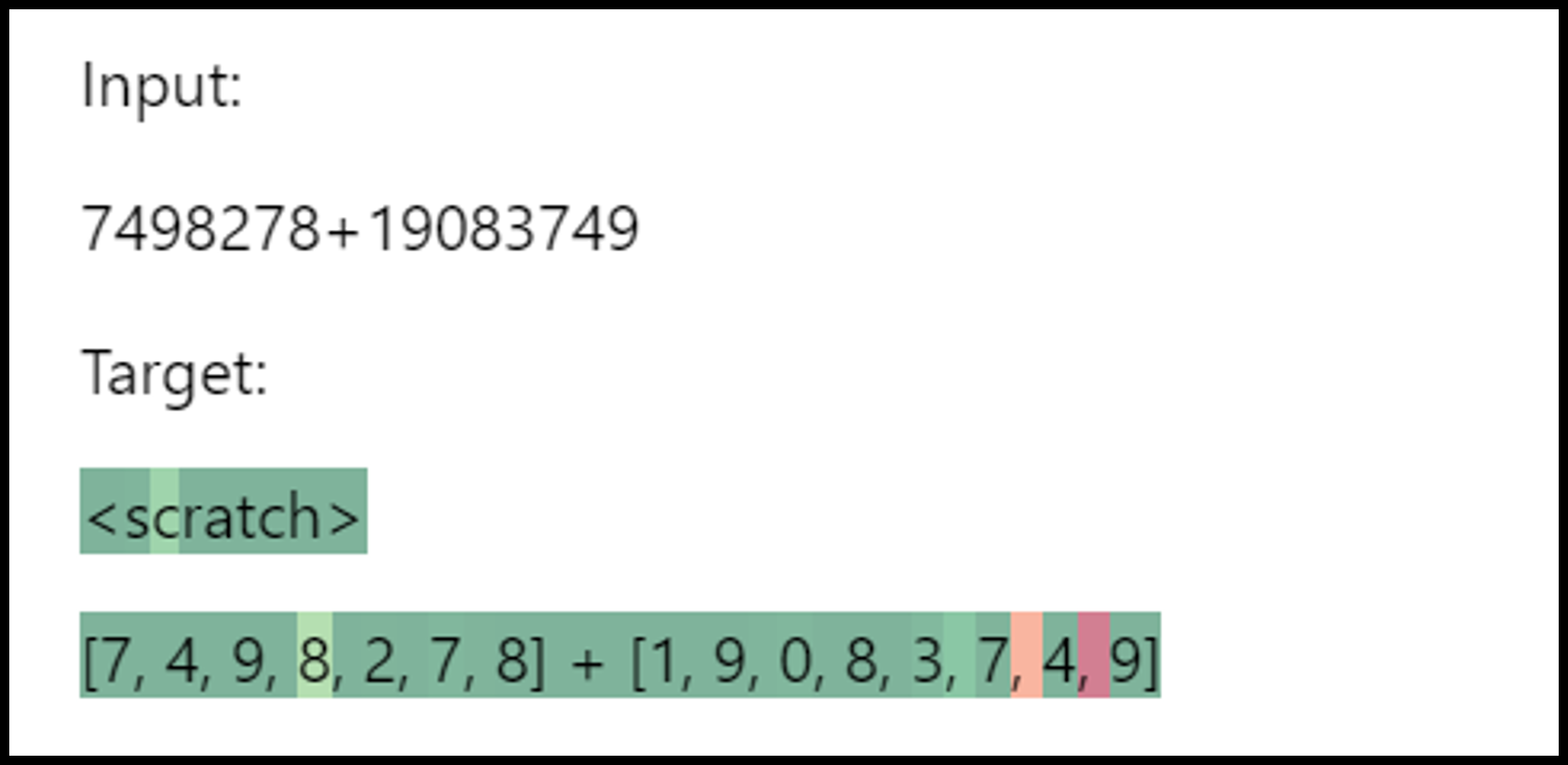}
        \caption{}
        \label{fig:ex_criticality_7}
    \end{subfigure}
     % \hfill
     \begin{subfigure}[b]{\columnwidth}
        \centering
        \includegraphics[width=\textwidth]{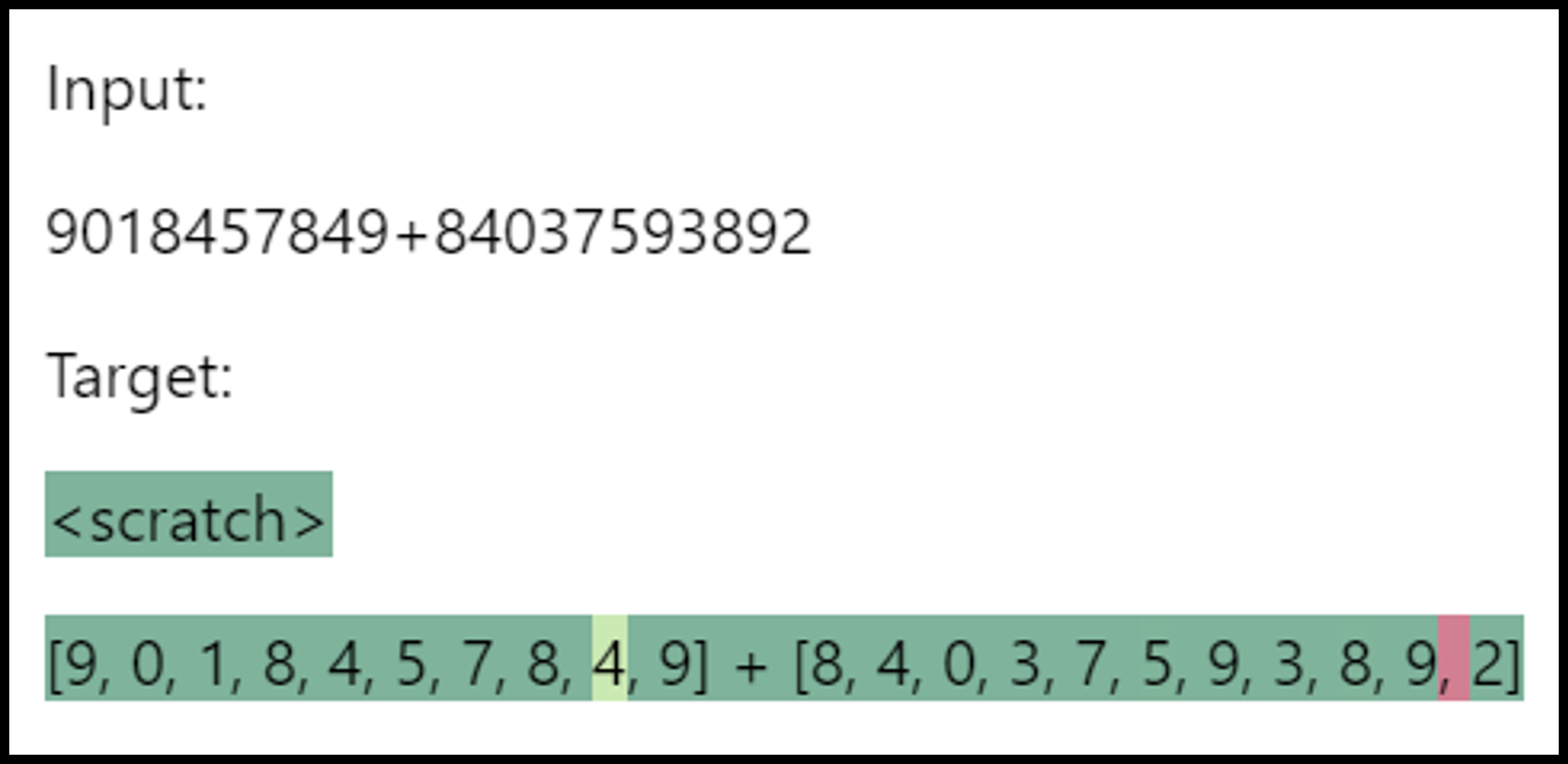}
        \caption{}
        \label{fig:ex_criticality_9}
    \end{subfigure}
        \caption{Output examples for addition tasks on $N+1$ digit lengths (the model is faced with numbers one notch longer than those encountered in pre-training). Each generated token is colored according to its certainty. A green color is a maximal certainty, while a red color is a minimal certainty.}
        \label{fig:examples_uncertainty}
\end{figure}

In this section, we provide insight into critical tokens, that play a crucial role in determining the success of the addition task.
Consider a pre-trained model on additions of numbers up to $N$ digits.
Now, consider a generalization test in which the model is prompted to add numbers with $N+1$ digits.
% Our experiments reveal that when the model fails at this task, the failure can typically be traced back to errors made on a very small number of tokens, which we refer to as critical tokens.
Our experiments reveal that when the model fails at this task, the failure can typically be traced back to errors made on critical tokens.
% These tokens are key decision points that completely determine the outcome of the addition process.
%Furthermore,
We observe that these critical tokens arise at the stage of the generation where the model must choose whether to treat the problem as an addition of $N$-digit numbers -- leading to failure -- or correctly addressing the task of adding $(N+1)$-digit numbers. 
More precisely, this error is caused by the omission of digits when copying the numbers from the previous step.
Figure~\ref{fig:ex_certainty} shows two examples of failed generation caused by errors on the critical tokens.
In the first case, the model pre-trained on numbers up to $3$ digits mistakenly recopies the last digit instead of the penultimate digit, leading to an incorrect outcome.
In the second example, where the model is pre-trained on numbers up to $5$ digits, it incorrectly closes the bracket in both cases instead of inserting a comma (the stage preceding the copying of the sixth digit).
These examples illustrate two types of critical tokens.
We only show them on the first decomposition line, but they can be found on the subsequent lines as well.
% For each operand, both types of errors can occur, meaning that for two numbers, there can be as many as four critical tokens per line.

% \begin{figure}
%     \centering
%     \includegraphics[width=\linewidth]{figures/criticallity/example_critical_token_3_5digits_1_uncolored.png}
%     \caption{Example of failed generations with an arrow to identify the critical token}
%     \label{fig:critical_tokens}
% \end{figure}

% \subsection{Criticality and Uncertainty}

% In this section, we provide both a qualitative and a quantitative rationale for introducing the prioritized KL penalty in Section~\ref{sec:weightedKLtrick}. \\

As explained in Section~\ref{sec:weightedKLtrick}, we quantify the certainty of model being in state $s$ through the quantity $\widehat{J}_{\theta_{\text{old}}}(s)$.
To provide more visual understanding of this quantity, we display in Figure~\ref{fig:examples_uncertainty} a few output examples with the colors as indication of the model certainty (green: high certainty, red: low certainty). % as in Figure~\ref{fig:critical_tokens} augmented with $\widehat{J}_{\theta_{\text{old}}}(s)$. 

\section{Assessing the impact of the certainty exponent \texorpdfstring{$\beta$}{Lg}}\label{app:ablation_study}

\begin{figure}
    \centering
    \includegraphics[width=0.99\linewidth]{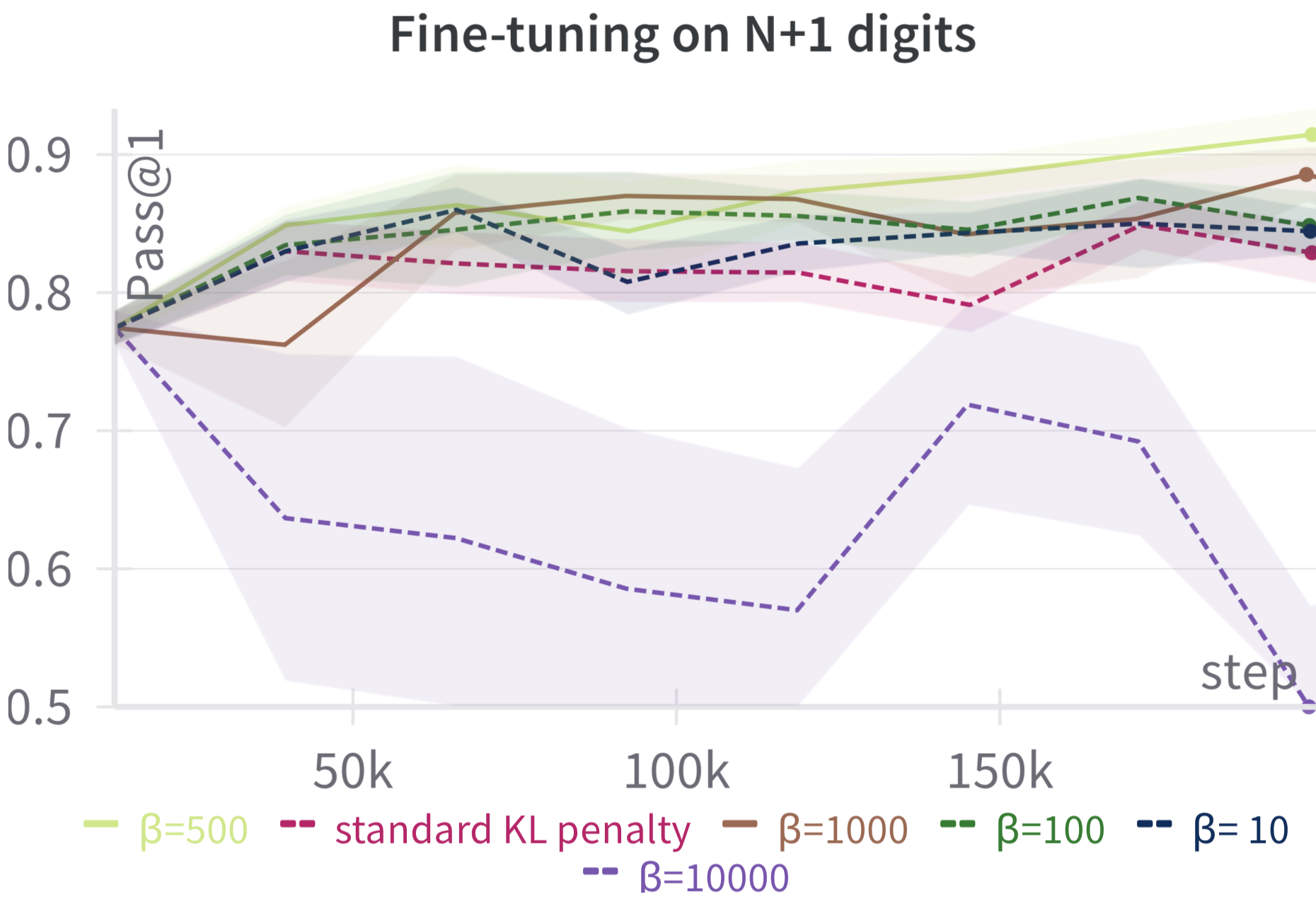}
    \caption{Fine-tuning results with various values of $\beta$ (averaged over 9 random seeds)}
    \label{fig:ablation_study_beta}
\end{figure}

In order to better assess the robustness of our prioritized KL penalty, we have carried out an experiment testing multiple orders of magnitude for the value of the $\beta$ exponent in Equation \ref{eq:weighted_kl}.
The corresponding learning curves are reported in Figure \ref{fig:ablation_study_beta}.
Despite important error margins, these results show that the prioritized KL penalty slightly outperforms the standard KL penalty for values of $\beta$ ranging from 10 to 500, reaching its maximum at $\beta=500$ and starting to decline from $\beta=1000$ (which shows early signs of instability).
The performance drops drastically at $\beta=10000$.
The good performance over such a wide range of beta values can be explained by the fact that after our pre-training, the confidence of the model is extremely high (except on critical tokens), which is why it takes large values of $\beta$ to drastically reduce the weight $\widehat{J}_{\theta_{\text{old}}}(s)^{\beta}$ in the prioritized KL penalty.
Therefore, we believe that this range (10-500) of acceptable $\beta$ values might be quite different in another problem.

\section{Experiments Details}\label{app:experiments_details}

The hyperparameters used in the experiment from Section~\ref{subsec:compare_pretrain} are provided in Table~\ref{table:hyperparams_compare_pretrain}.
The hyperparameters used in the experiment from Section~\ref{subsec:compare_kl_trick} are provided in Table~\ref{table:hyperparams_compare_kl_trick}.

\begin{table}[ht]
    \centering
    \medskip
    \begin{tabular}{ccc}
        \toprule
        Hyperparameter & Value\\ \midrule
        Learning rate                          & $10^{-6}$  \\
        Discount factor                        & $1$ \\
        Value function coefficient             & $0.1$ \\
        Entropy coefficient                    & $0.0005$ \\
        KL penalty coefficient                 & $10$ \\
        Repeat per collect                     &  $1$  \\
        Episodes per collect                   &  $50$  \\
        Episodes per test                      & $100$\\
        \bottomrule
    \end{tabular}
    \caption{Hyperparameters used in the RL experiment comparing multiple levels of pre-training}
    \label{table:hyperparams_compare_pretrain}
\end{table}

\begin{table}[ht]
    \centering
    \medskip
    \begin{tabular}{ccc}
        \toprule
        Hyperparameter & Value\\ \midrule
        Learning rate                          & $10^{-6}$  \\
        Discount factor                        & $1$ \\
        Value function coefficient             & $0.1$ \\
        Entropy coefficient                    & $0.0005$ \\
        KL penalty coefficient                 & $5$ \\
        KL penalty exponent ($\beta$)            & $150$ \\
        Repeat per collect                     &  $1$  \\
        Episodes per collect                   &  $50$  \\
        Episodes per test                      & $100$\\
        \bottomrule
    \end{tabular}
    \caption{Hyperparameters used in the RL experiment evaluating the impact of the prioritized KL penalty}
    \label{table:hyperparams_compare_kl_trick}
\end{table}

\subsection{Detailed Pretraining Results}\label{app:detailed-pretraining-results}

\begin{table*}
    \centering
    \scalebox{0.95}{
    \begin{tabular}{c|cccc}
        \toprule
        \makecell{Nb. of \\ Digits}  & \makecell{$N$\\ Accuracy} & \makecell{$N+1$\\ Accuracy} & \makecell{$N+2$\\ Accuracy} & \makecell{$N+3$\\ Accuracy} \\ \midrule
        7  & 98.9\% $\pm$ 0.7\% & 48.8\% $\pm$ 3.0\% & 0.0\% $\pm$ 0.0\% & 0.0\% $\pm$ 0.0\% \\
        9  & 96.4\% $\pm$ 0.6\% & 78.9\% $\pm$ 2.4\% & 0.5\% $\pm$ 0.5\% & 0.0\% $\pm$ 0.0\% \\
        11 & 91.2\% $\pm$ 1.3\% & 75.1\% $\pm$ 2.7\% & 30.7\% $\pm$ 2.4\% & 0.2\% $\pm$ 0.3\% \\
        13 & 93.0\% $\pm$ 1.6\% & 88.9\% $\pm$ 2.1\% & 67.7\% $\pm$ 3.1\% & 20.4\% $\pm$ 2.4\% \\
        \bottomrule
    \end{tabular}}
    \caption{Model accuracy on addition tasks with identical digit lengths.}
    \label{tab:accuracy1}
\end{table*}

\begin{table*}
    \centering
    \scalebox{0.95}{
    \begin{tabular}{c|cccc}
        \toprule
        \makecell{Nb. of \\ Digits}  & \makecell{$N$\\ Accuracy} & \makecell{$N+1$\\ Accuracy} & \makecell{$N+2$\\ Accuracy} & \makecell{$N+3$\\ Accuracy} \\ \midrule
        7  & 100.0\% $\pm$ 0.0\% & 69.0\% $\pm$ 2.4\% & 0.0\% $\pm$ 0.0\% & 0.0\% $\pm$ 0.0\% \\
        9  & 97.0\% $\pm$ 0.6\% & 89.4\% $\pm$ 1.8\% & 6.9\% $\pm$ 1.3\% & 0.0\% $\pm$ 0.0\% \\
        11 & 94.4\% $\pm$ 1.4\% & 87.0\% $\pm$ 2.1\% & 53.7\% $\pm$ 3.2\% & 7.3\% $\pm$ 1.6\% \\
        13 & 95.6\% $\pm$ 1.4\% & 92.5\% $\pm$ 1.9\% & 84.7\% $\pm$ 2.4\% & 51.8\% $\pm$ 3.2\% \\
        \bottomrule
    \end{tabular}}
    \caption{Model accuracy on addition tasks with varying digit lengths.}
    \label{tab:accuracy2}
\end{table*}

Tables \ref{tab:accuracy1} and \ref{tab:accuracy2} display the model's performance on addition tasks for different digit lengths that the model was pretrained on. These digit lengths refer to the number of digits used during pretraining (7, 9, 11, and 13 digits), with accuracy then measured on tasks involving identical digit lengths and varying digit lengths. The model is subsequently evaluated on its ability to generalize to more complex tasks, i.e., $N+1$, $N+2$, and $N+3$ digits, where the total number of digits exceeds the training range.

Across both tables, the general trend indicates that the model is more adept at solving tasks within its training range, and it exhibits improved generalization with larger digit lengths training. However, in both identical and varying digit tasks, the model's ability to handle tasks involving $N+2$ and $N+3$ is limited, particularly for smaller digit lengths. This suggests that while pretraining enables the model to generalize to some extent, there are clear limitations when the task complexity surpasses the data on which the model was trained.

\subsection{Details on the fine-tuning with RL experiments}\label{app:rl_experiments}

\begin{figure*}
     \centering
     \begin{subfigure}[b]{0.49\textwidth}
         \centering
         \includegraphics[width=\linewidth]{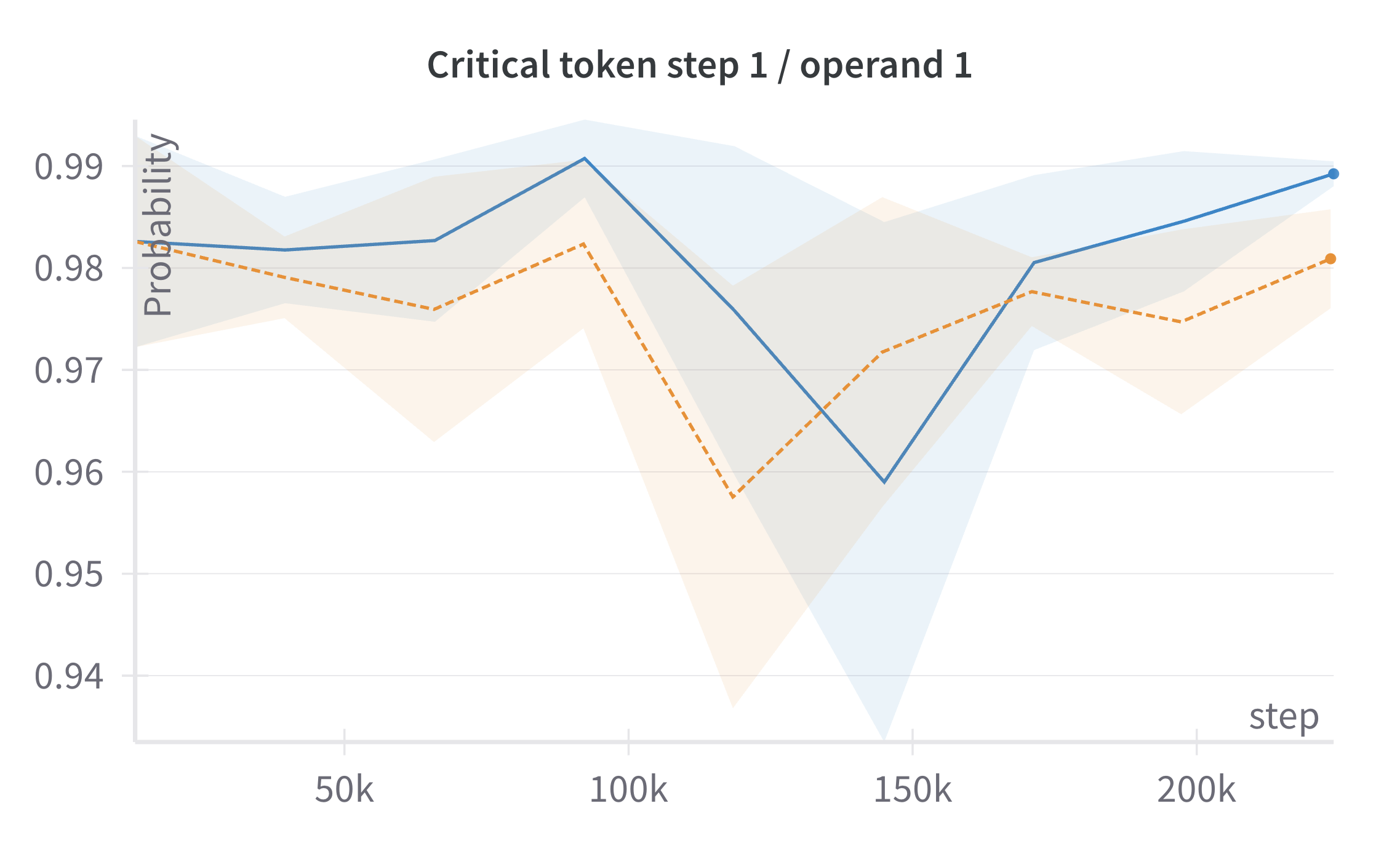}
     \end{subfigure}
     \hfill
     \begin{subfigure}[b]{0.49\textwidth}
         \centering
         \includegraphics[width=\linewidth]{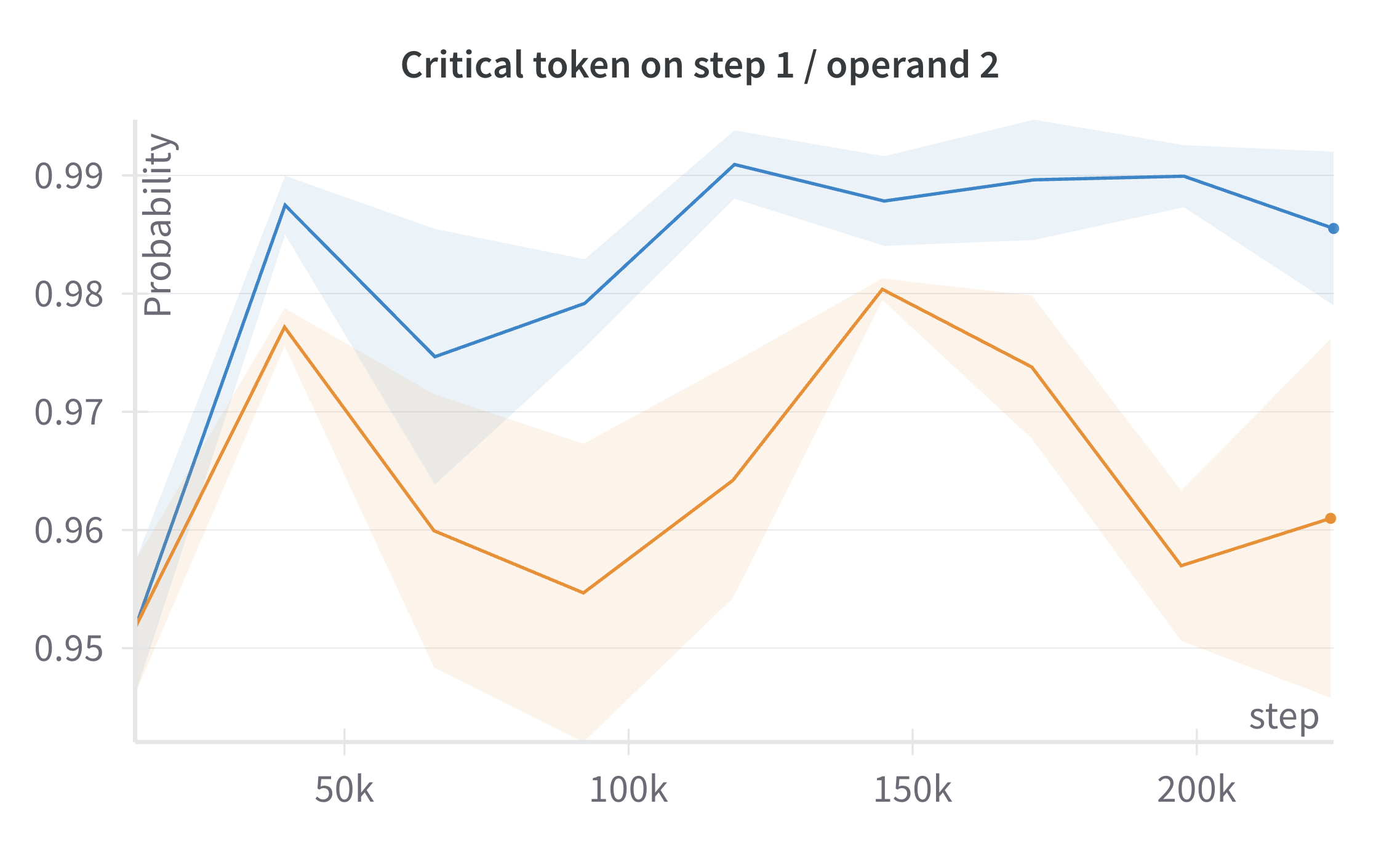}
     \end{subfigure}
     \vspace{1em} % Space between rows
     \begin{subfigure}[b]{0.49\textwidth}
         \centering
         \includegraphics[width=\linewidth]{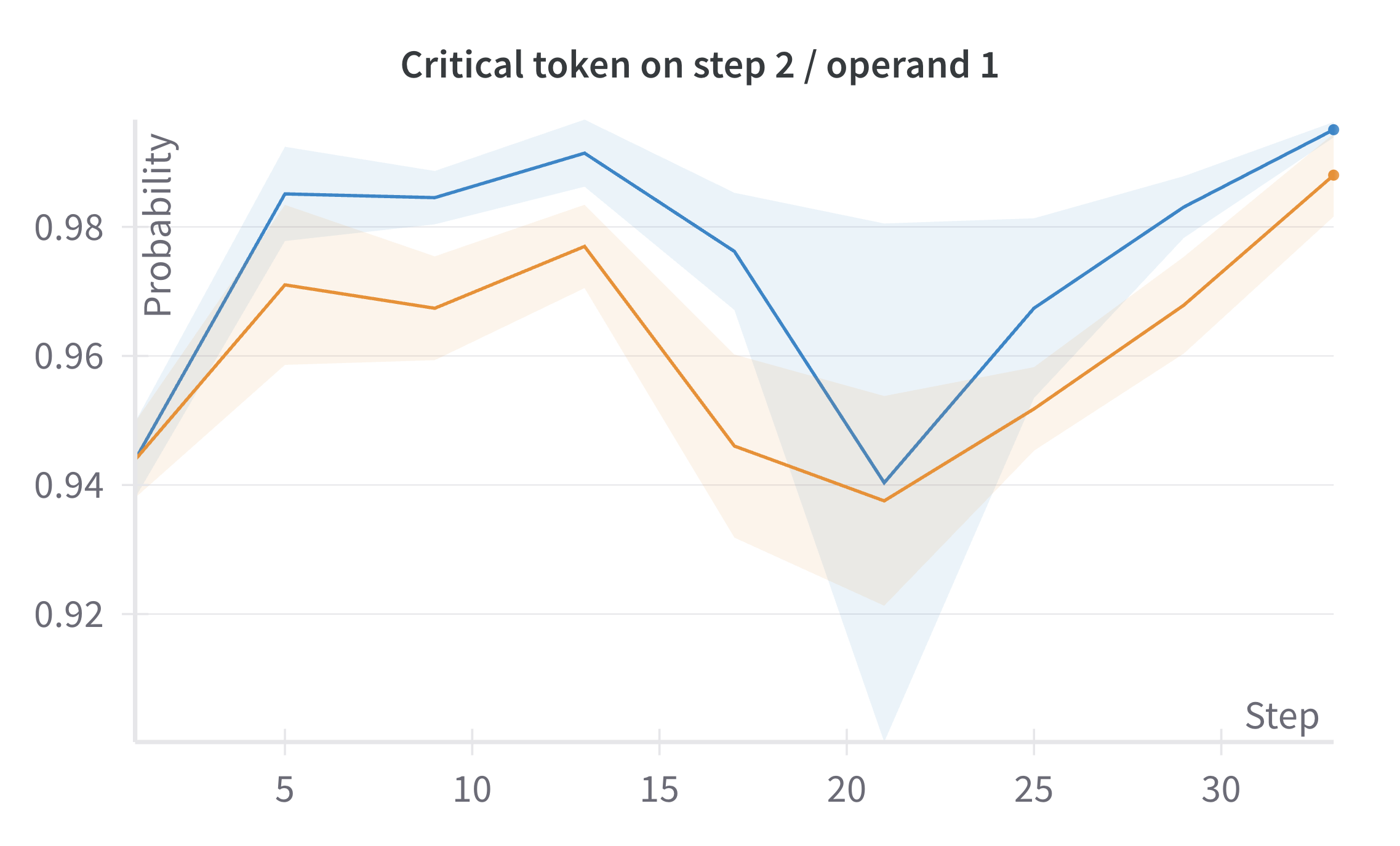}
     \end{subfigure}
     \hfill
     \begin{subfigure}[b]{0.49\textwidth}
         \centering
         \includegraphics[width=\linewidth]{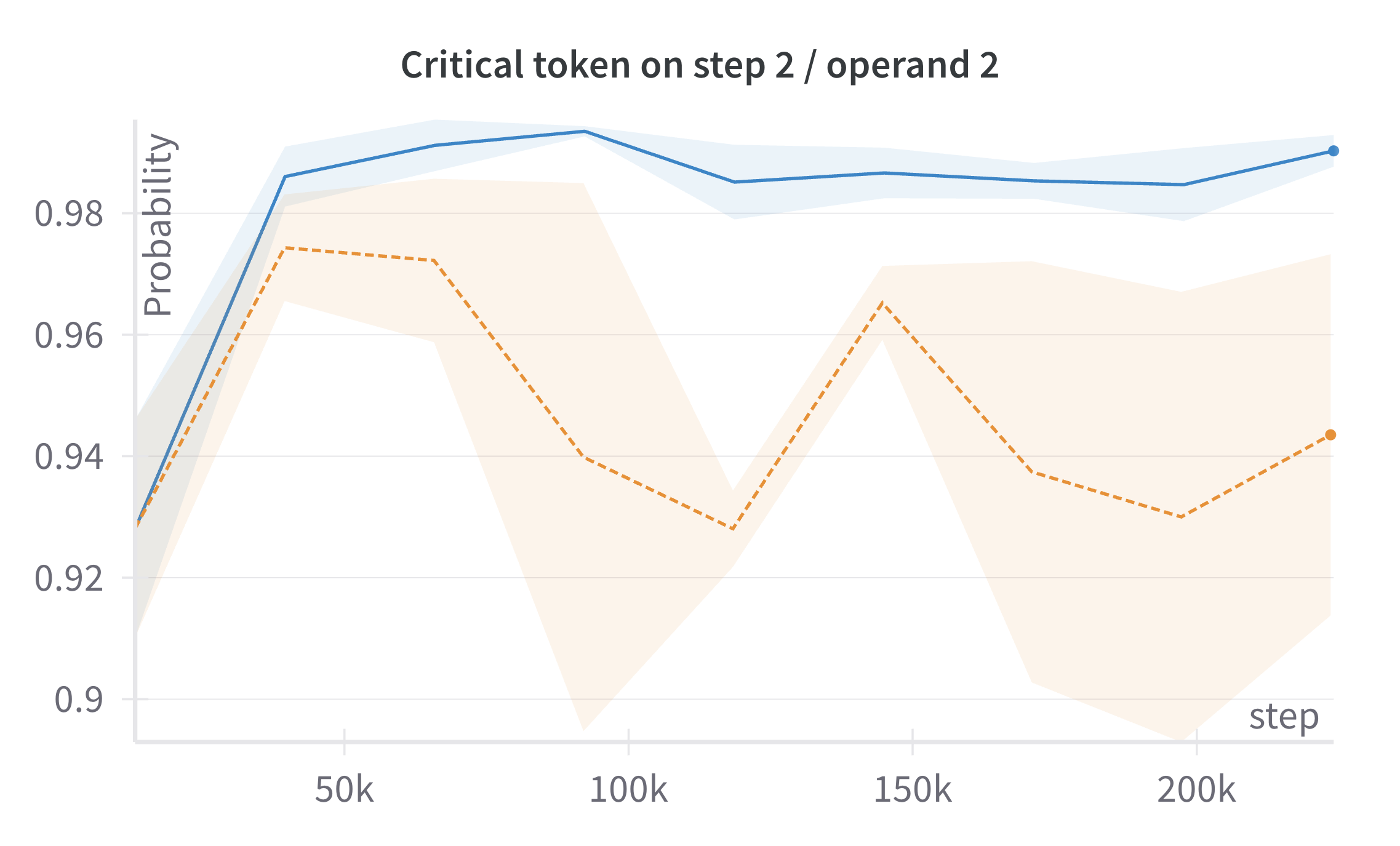}
     \end{subfigure}
     \vspace{1em} % Space between rows
     \begin{subfigure}[b]{0.49\textwidth}
         \centering
         \includegraphics[width=\linewidth]{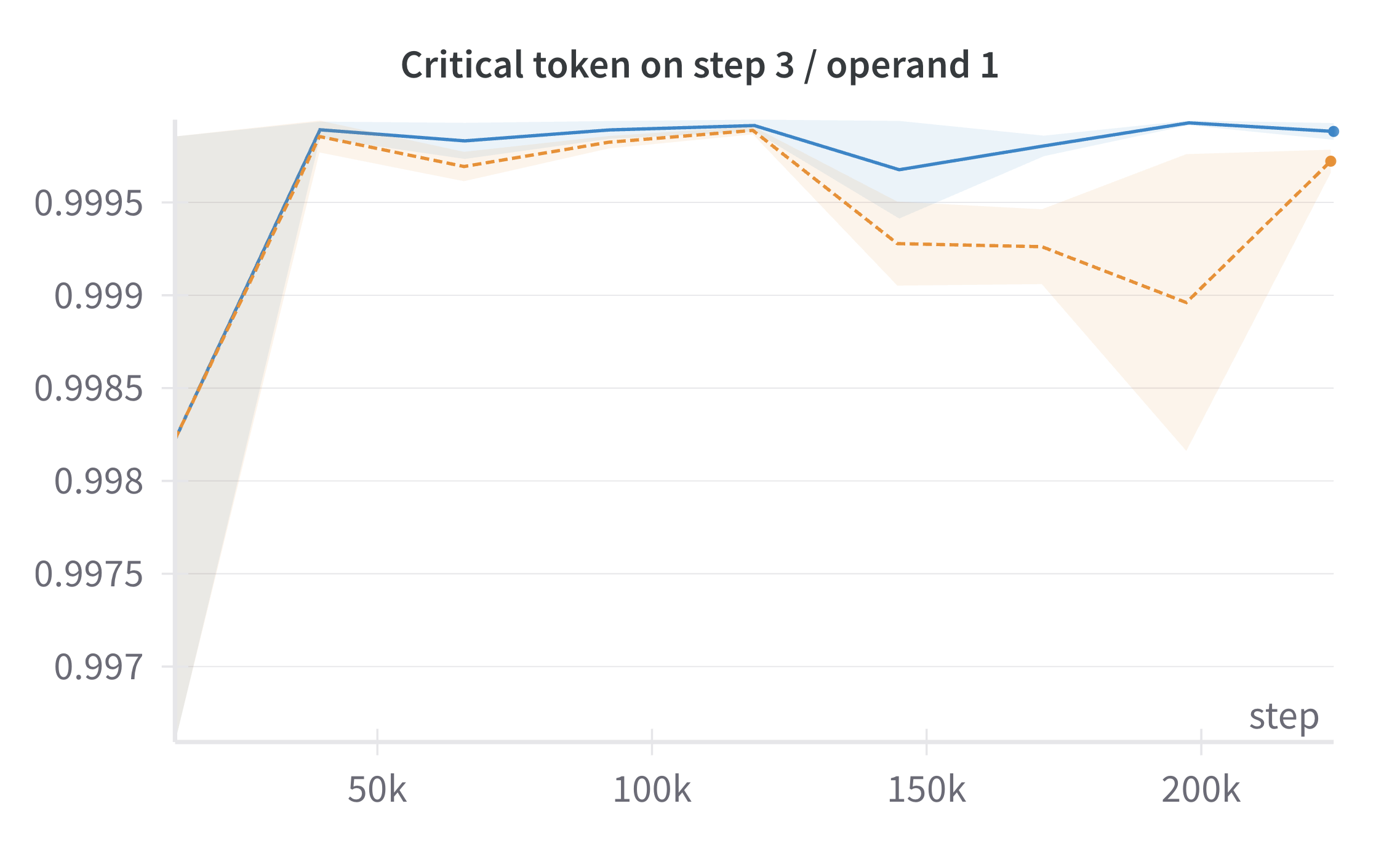}
     \end{subfigure}
     \hfill
     \begin{subfigure}[b]{0.49\textwidth}
         \centering
         \includegraphics[width=\linewidth]{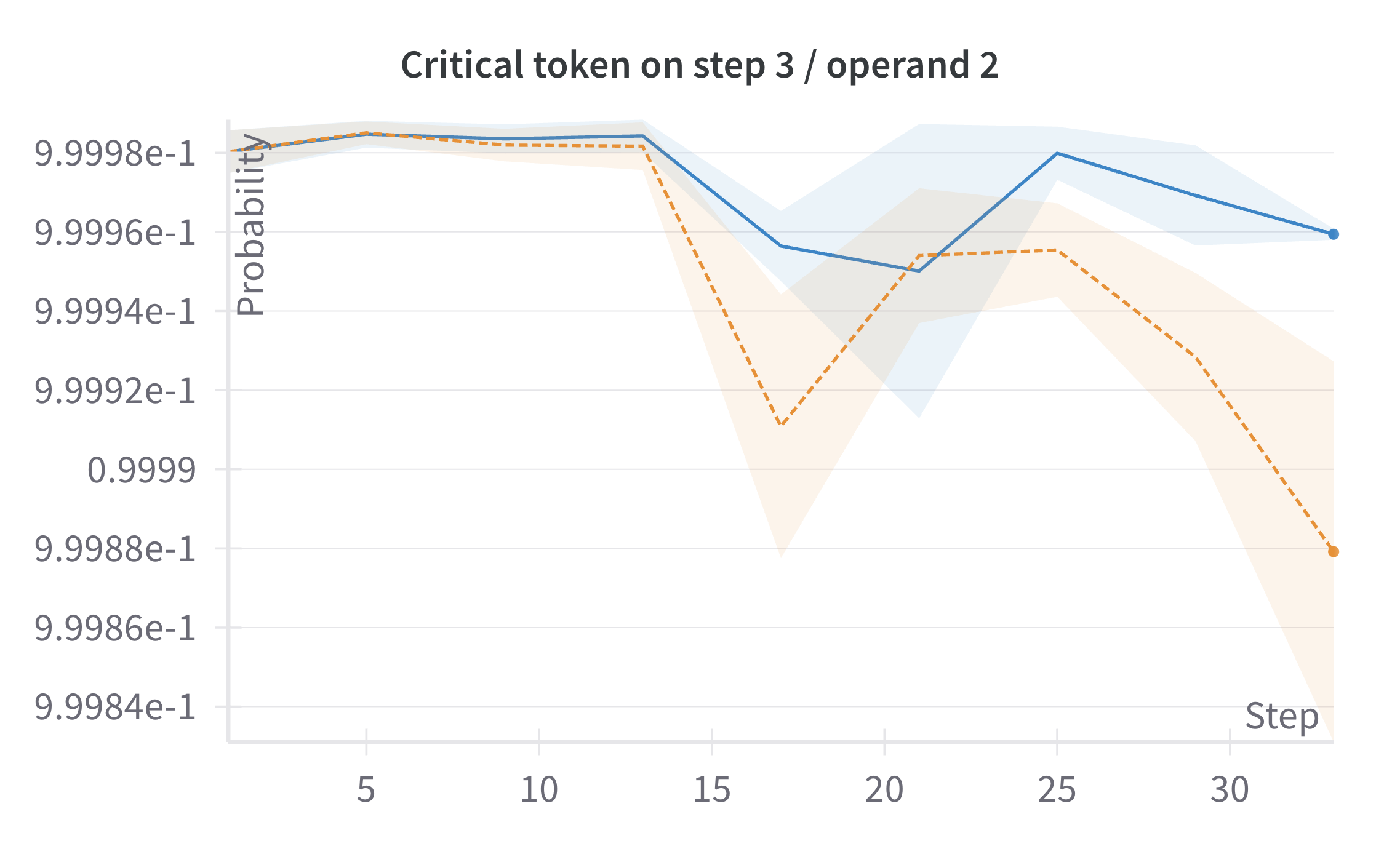}
     \end{subfigure}
     \begin{subfigure}[b]{0.49\textwidth}
         \centering
         \includegraphics[width=\linewidth]{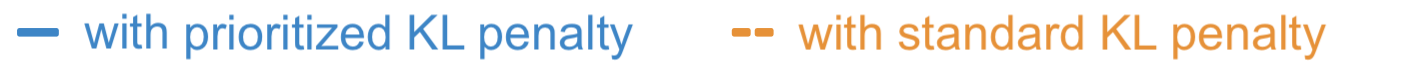}
         \vspace{1em} % Space between rows
     \end{subfigure}
        \caption{Evolution of the right prediction probability on multiple critical tokens, during the RL fine-tuning on number length $N+1=8$.}
    \label{fig:ct_probas}
\end{figure*}

In Figure \ref{fig:ct_probas} we expose the evolution of the right prediction probability for 6 different critical tokens.
These critical tokens are selected as the commas on the $(N+1)$-th token for each operand list, which is a frequent source of errors.
One can observe that in each situation (despite important error margins), the probabilities outputted by the model trained with prioritized KL penalty are higher than the other.
Note that this effect is more pronounced on tokens ``step 1 / operand 2'' and ``step 2 / operand 2'' as on the others, the probability of success is already very high from the start.

\section{Softwares}

We carried out our experiments using the Python packages Transformers \cite{wolf-etal-2020-transformers} and Tianshou \cite{tianshou}.

\end{document}